# IPDAE: Improved Patch-Based Deep Autoencoder for Lossy Point Cloud Geometry Compression

Kang You
Nanjing University of Aeronautics and Astronautics
Nanjing, China
youkang@nuaa.edu.cn

Pan Gao
Nanjing University of Aeronautics and Astronautics
Nanjing, China
pan.gao@nuaa.edu.cn

Qing Li
Southern University of Science and Technology
Shenzhen, China
liq36@sustech.edu

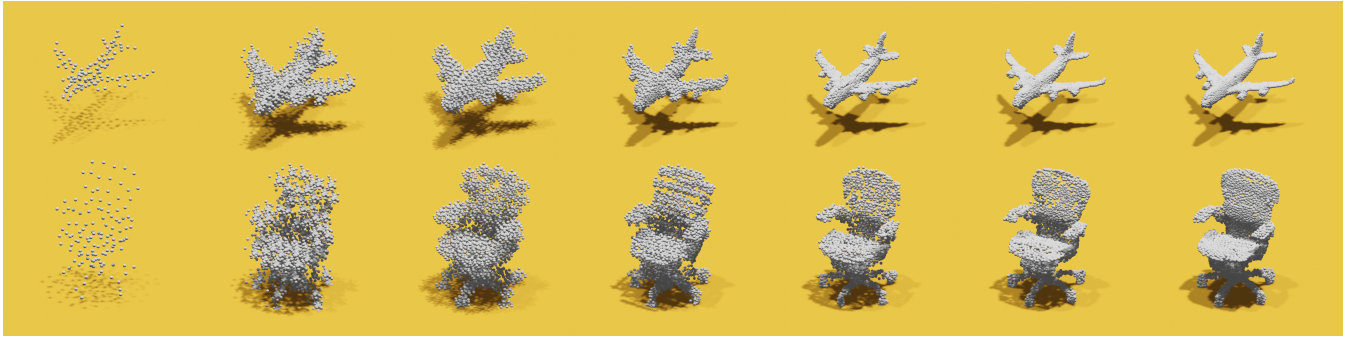

Figure 1: Visualization of Training Process of the Proposed IPDAE in This Paper.

## ABSTRACT

Point cloud is a crucial representation of 3D contents, which has been widely used in many areas such as virtual reality, mixed reality, autonomous driving, etc. With the boost of the number of points in the data, how to efficiently compress point cloud becomes a challenging problem. In this paper, we propose a set of significant improvements to patch-based point cloud compression, i.e., a learnable context model for entropy coding, octree coding for sampling centroid points, and an integrated compression and training process. In addition, we propose an adversarial network to improve the uniformity of points during reconstruction. Our experiments show that the improved patch-based autoencoder outperforms the state-of-the-art in terms of rate-distortion performance, on both sparse and large-scale point clouds. More importantly, our method can maintain a short compression time while ensuring the reconstruction quality.

## CCS CONCEPTS

• **Computing methodologies** → **Point-based models**; *Reconstruction*; Image compression.

## KEYWORDS

point cloud geometry, lossy compression, deep learning, adversarial learning

## 1 INTRODUCTION

The 3D point cloud is an important representation of objects or scenes in 3D space [1, 7, 10, 34]. The highly challenging task of compressing the shape information represented by a set of coordinates of points in a point cloud is referred to as point cloud geometry (PCG) compression [11, 25, 27, 35]. With the rapid development of point cloud applications in recent years, PCG compression is gaining more and more attention.

Compared to image/video compression [5, 15, 18, 19], point cloud compression poses three major challenges. Firstly, a point cloud is an unordered set of vectors, which does not have regular structure. This requires that the compression model can handle irregularity and consume unordered input sets. Secondly, the spatial resolution of point clouds may vary hugely, ranging from thousands of points to millions of points. This requires the compression model generic, which can compress both small-scale and large-scale point cloud time-efficiently. Thirdly, due to lack of connectivity in the point cloud, the reconstructed point clouds may clutter together. Having said that, the generated output point cloud should be more uniform.

At present, the methods for PCG compression can be classified into three categories: traditional compression algorithm, 3D convolution based autoencoder, and PointNet based autoencoder. The first one is model-driven method, which addresses point cloud compression by using octree or hand-crafted features. The latter two belong to data-driven approaches. While 3D convolution based autoencoders need the transformation of points to regular 3D voxels, rendering data unnecessarily voluminous, the PointNet based solution seems efficient and effective. In [32], the authors divide the point cloud into patches for training and compression, which improves the performance of PointNet based autoencoder and achieves state-of-the-art performance. However, the autoencoder does not consider the global point cloud information, and only uses the optimization of local reconstruction to approximate the optimization of global reconstruction. This ignores the correlation between patches, which will cause information redundancy during compression. Thus, the developed autoencoder cannot achieve optimal reconstruction shape, especially for large-scale point cloud. In addition, the uniformity of reconstructed point cloud is not considered.

In this paper, to meet above challenges and issues, we propose an end-to-end deep point cloud compression model, which makes a series of significant improvements on patch-based point cloud autoencoder. Specifically, the contributions of this paper include:

1) Learnable context model for entropy coding. We use sampling centroid points as the context information to estimate the entropy model of hidden representation of each patch, i.e., the probability distribution of each value of each patch's bottleneck layer. To reduce bit consumption, the sampling points are compressed using octree coding.

2) Integrated compression and training process. Different from the method of using the training set of independent patches to train the autoencoder in [32], we improve the training process to facilitate the comparison of distortion and bitrate between the input and output point clouds using global optimization criteria.

3) Adversarial learning. We design a discriminator and use the method of adversarial learning to make the reconstruction of autoencoder more uniform. We propose a metric to evaluate the uniformity of reconstructed point clouds. By using previously trained autoencoder weights as a starting point for the generator, we can reconstruct a more uniform point cloud.

## 2 RELATED WORK

With the increasing capability of 3D acquisition device, it becomes critical that how to efficiently compress 3D point clouds. Currently, PCG compression methods are mainly focused on octree based solutions [8, 14, 16, 24, 35], 3D convolutional autoencoder [22, 23, 27, 28] and PointNet-based autoencoder [29, 31, 32].

Geometry-based Point Cloud Compression (G-PCC) [16] is a compression standard recently developed by MPEG organization, which contains compression algorithms based on octree and other nested partitions. Taking octree as an example, the algorithm recursively divides the point cloud coordinate space into an octree. This algorithm is easy to understand and implement, but it cannot perform well at low bit rate, and the number of points generated will decrease sharply with the decrease of tree depth. The authors of [35] uses a region-wise processing to extract information redundancy between point cloud surface regions, which can reduce the number of bits required for compression to a certain extent. Some methods also combine deep learning with octree method to pursue better performance, such as [8, 24].

Quach *et al.* used a 3D autoencoder to compress point cloud geometry data in [22]. They preprocessed the point cloud with voxelization, and performed the process of analysis and synthesis transforms using a series of 3D convolution operations. As an improved version of 3D autoencoder, the authors of [23] proposed a set of contributions to improve deep point cloud compression, including entropy modeling, deeper transforms, etc. Although this method has achieved good results on some dense voxelized point clouds, it cannot perform well on sparse or irregular point clouds, which are more common in practical applications.

As the most representative voxel-based method at present, PCGC-v2 [27] uses a progressive re-sampling autoencoder to compress point clouds. Specifically, they extract features from the downsampled point cloud level-by-level, and then hierarchically reconstruct the original point cloud from sparse sampled points. It also uses sparse convolutions [6] to reduce compression time. Like other voxel-based methods, this method mainly aims at compression of dense voxelized point clouds, and may have significant performance drop on the more general point cloud, i.e., non-voxlized point clouds.

By directly taking the original points as input, PointNet [20] uses a symmetric function to extract features from the coordinate set. [31] designed a PointNet-based autoencoder using PointNet as the encoder and multilayer perceptron as the decoder. It can handle point cloud compression for simple shapes in the ShapeNet dataset, but not for relatively complex point clouds such as ModelNet40. PointNet++ [21] is a deep point cloud learning model developed on top of PointNet, and it uses multiple local PointNet operations to capture details of a point cloud shape. [12] designed a hierarchical structure similar to PointNet++ for geometry compression, but it can only be used for compression of sparse point clouds with very low resolution. Recently, the authors of [32] proposed a patch-based autoencoder for lossy geometry compression, which demonstrates superiority in compression on small benchmarks, such as ShapeNet and ModelNet40. In this paper, we conduct a series of methodological improvements to this approach, including learnable context model, training the compression model at the sample level with all patches involved, and adversarial learning. Our improved autoencoder can generalize well to large-scale datasets.

## 3 PROPOSED COMPRESSION STRATEGY

In this section, we will demonstrate the set of contributions we proposed for lossy geometry coding. Figure 2 shows our entire end-to-end compression architecture. Figure 4 and 5 are specific descriptions of the corresponding simplified parts in Fig. 2.

### 3.1 Integrated compression and training process

In the recent work [32], the authors train the point cloud autoencoder patch-by-patch, and then use the pre-trained autoencoder to compress the whole point cloud. In this work, we integrate the compression and training process together. That is, we train the autoencoder in a point-cloud by point-cloud manner, and use the loss for the point cloud sample with all patches involved to optimize the network. Then, we employ the trained model to directly compress the point cloud.

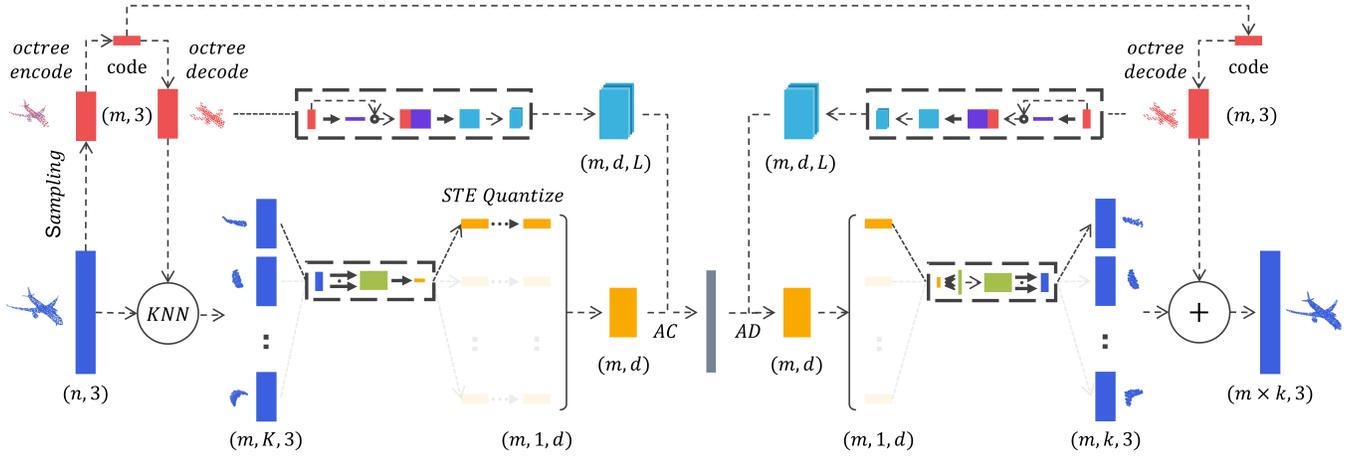

**Figure 2: Improved end-to-end point cloud compression network architecture. The model takes a point cloud of dimension of $(n, 3)$ as input, and reconstructs a point cloud of dimension of $(m \times k, 3)$. $m$ denotes the number of patches we divide, and $K$ and $k$ are the point counts in the encoded and decoded patches, respectively. $d$ is the feature dimension of patch embedding, and $L$ is the quantization level. In training, we replace arithmetic coding and arithmetic decoding with bit rate estimation.**

At the beginning, we combine octree with FPS and KNN to propose a scheme for orderly dividing a point cloud into patches. Compared to previous work directly transmitting the centroid points without any compression, this scheme can effectively compress the sampling point information of the point cloud during entropy coding, saving a number of bit rates. Further, through this method, we can get a group of patches sorted according to the spatial structure of octree partition, which is beneficial for reconstructing the whole point cloud.

Let $X$ denote a shape in three-dimensional space, which is a three-dimensional random variable. Then, a point cloud containing $n$ points is a sample of capacity $n$ from the population $X$, and it can be written as $S = \{x_1, x_2, \ldots, x_n\}$.

After applying the farthest point sampling for $S$, we can obtain the results $S' = \{x'_1, x'_2, \ldots, x'_m\}$. We firstly use octree to entropy encode the sampling points $S'$ to transmit the sampling points at a lower bit rate; meanwhile, in order to avoid drift error, we also do decoding for $S'$ at the encoder. For each decoded sampling point $x'_{(i)}$, we use KNN to find its neighboring points $\{x^1_i, \cdots, x^K_i\}$. The relative positions of neighboring points with respect to the centriod point form a patch of $S$, which can be represented as $S_{(i)} = \{x^1_i - x'_{(i)}, x^2_i - x'_{(i)}, \ldots, x^K_i - x'_{(i)}\}$. Each patch is then gone through an autoencoder for compression. The detail of the proposed patch division scheme is presented in Fig. 3. At the decoder, we firstly get the sampling centroid points by using octree decoding. The decoded centroid points can be represetned as $\widetilde{S'} = \{x'_{(1)}, x'_{(2)}, \ldots, x'_{(m)}\}$, where $x'_{(m)}$ denotes the reconstructed position of centriod point $m$. For each patch, the autodecoder result can be represented as $\widetilde{S}_{(i)} = \{(x^1_i - x'_{(i)})', (x^2_i - x'_{(i)})', \ldots, (x^K_i - x'_{(i)})'\}$, where $(x^K_i - x'_{(i)})'$ represents the reconstructed relative position of the $K$ neighbouring point in patch $i$. Combining with the results of octree decoding and autodecoder, we can reconstruct the $i$th patch by using $\hat{S}_{(i)} = \{(x^1_i - x'_{(i)})' + x'_{(i)}, (x^2_i - x'_{(i)})' + x'_{(i)}, \ldots, (x^K_i - x'_{(i)})' + x'_{(i)}\}$, and finally obtain the whole point cloud $\hat{S}$.

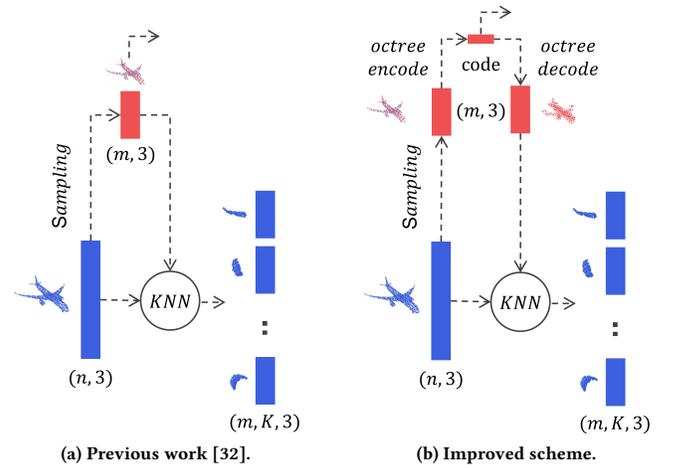

(a) Previous work [32].    (b) Improved scheme.

**Figure 3: Patch-division comparison, where we incorporate an octree encoding and decoding process.**

After having obtained the patches, we normalize the scale of each input point cloud as a preprocessing step. Specifically, in the experiment, we move the center point of each point cloud to $(0.5, 0.5, 0.5)$, and then scale its coordinates to $[0, 1]$. In the compression test, after we get the final output result, we also transform it to the original center point and do scaling accordingly. This can ensure the uniformity of point cloud data distribution to the greatest extent and make the training results exhibit better performance. At the same

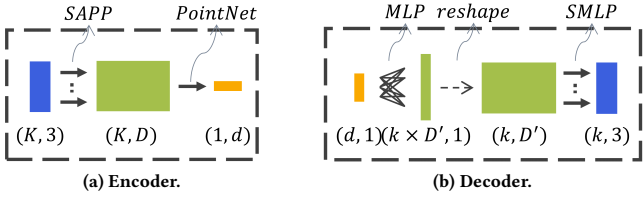

(a) Encoder.    (b) Decoder.

Figure 4: Improved autoencoder detail, where SMLP and SAPP refer to shared MLP and set abstraction per point, respectively. For encoder, $D$ is the output dimension of an MLP. For decoder, $k$ is the point count of a generated patch, and $D'$ is the feature dimension.

time, it is also convenient to compress point clouds in different coordinate intervals.

It is noteworthy that when performing octree coding on the sampled structure points, we need to achieve the balance between reconstruction quality and bit rate. We can achieve this either by fixing the depth of the octree or by directly setting the bit rate of octree coded bitstream to a certain value. In our experiments, we use the latter to make the compression process more intuitive.

The autoencoder structure we used in this paper is shown in the Fig. 4, where we consider a deep decoder to improve the decoding capability. In addition, we must take into account that the densities of the points, since the patches obtained from dense and sparse point clouds could be very different, which can affect the performance of PointNet-based autoencoder. Hence, we set

$$S_{norm(i)} = \{ \sqrt[3]{\frac{n}{n_0}} \cdot x | x \in S_{(i)} \} \quad (1)$$

to reduce the effect of point density on data distribution to some extent. $n$ denotes the number of points contained in the input point cloud, $n_0$ denotes a reference quantity constant, and we set $n_0$ to 1024 in our experiments. After obtaining the output of the decoder, we perform the corresponding inverse transform on each reconstructed patch.

With an entropy model across all patches, we can directly estimate the bit rate of an entire point cloud and use Chamfer Distance $D_{cd}$ of a complete shape as the distortion metric. Using $Loss = D_{cd} + \lambda R$, we can carry out a complete end-to-end training for the compression architecture. The Chamfer Distance $D_{cd}$ is defined as follows

$$D_{cd}(S, \hat{S}) = \frac{1}{|S|} \sum_{x \in S} \min_{\hat{x} \in \hat{S}} \|x - \hat{x}\|_2^2 + \frac{1}{|\hat{S}|} \sum_{\hat{x} \in \hat{S}} \min_{x \in S} \|\hat{x} - x\|_2^2 \quad (2)$$

where $x$ is a point in the original point set $S$, and $\hat{x}$ denotes a point from the reconstructed point set $\hat{S}$. In the global loss for a point cloud, we set $\lambda$ to $10^{-6}$. The entropy model and the definition of bitrate $R$ will be presented in next section.

### 3.2 Learnable context model for entropy coding

After encoding each patch using an autoencoder, we have the latent representation of point cloud $y$, $y \in \mathbb{R}^{m \times d}$, where $d$ denotes the feature dimension of each patch embedding. We consider a straight-through estimator for quantization similar to [17], which can better handle the quantization of the latent representation. Define $y_{(i)j}$ as the $j$th element in the hidden representation of the $(i)$th patch. Given quantization level $L$ represented by a positive odd number, we can quantize each latent element as follows:

$$\hat{y}_{(i)j} = round\left(\sigma\left(y_{(i)j}\right) \times L - \frac{L}{2}\right) + \frac{L}{2} \quad (3)$$

where $1 \leq i \leq m$, $1 \leq j \leq d$ and $\sigma$ represents the *sigmoid* function. Equation 3 shows how we get the quantized value of $y$, and each quantized value $\hat{y}_{(i)j} \in \{0, 1, \ldots, L-1\} \subset Z$.

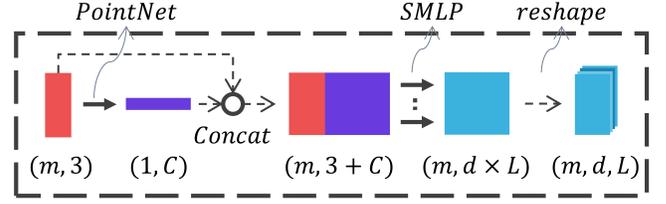

Figure 5: The proposed architecture for entropy modeling, where $C$ is the dimension of the output vector of the PointNet.

As for back propagation during training, we use gradients of the identity function:

$$\nabla \hat{y}_{(i)j} = \nabla(\sigma\left(y_{(i)j}\right) \times L) \quad (4)$$

For bit rate estimation, we use a neural network to predict the probability mass functions $p_{(i)j}$ of each element $\hat{y}_{(i)j}$ in the hidden representation based on the sampling points. The details of our network structure for probability modeling are shown in Fig. 5. Since each $y_{(i)j}$ has $L$ possible quantized values, the dimension of the output of the network is $(m \times d \times L)$. We use the *softmax* function for the last dimension of the output to ensure:

$$\sum_{\hat{y}_{(i)j}=0}^{L-1} p_{(i)j}(\hat{y}_{(i)j}) = 1 \quad (5)$$

Then, we can get the distribution for each element $\hat{y}_{(i)j}$ as follows:

$$\begin{array}{ccccc} \hat{y}_{(i)j} & 0 & 1 & \ldots & L-1 \\ p_{(i)j} & p_{(i)j}(0) & p_{(i)j}(1) & \ldots & p_{(i)j}(L-1) \end{array} \quad (6)$$

The bit rate estimation expression for latent representation of patches is given as follows:

$$R = \frac{1}{n} \sum_{i=1}^{m} \sum_{j=1}^{d} (-\log_2 p_{(i)j}(\hat{y}_{(i)j})) \quad (7)$$

where $p_{(i)j}(\hat{y}_{(i)j})$ refers to the probability of the actual value of quantized element $\hat{y}_{(i)j}$. In testing, we use the predicted probability mass function $p_{(i)j}$ to do arithmetic encoding and decoding.

### 3.3 Adversarial learning

In this section, we incorporate adversarial learning into our compression model. We expect that the proposed compression model combined with a discriminator encourages decompressed point cloud perceptually hard to distinguish from orignal point cloud. To this end, by using a pretrained autoencoder as a generator, we designed a new adversarial learning based model. Specifically, we use Wasserstein GAN (WGAN) [2, 9] to improve the stability of learning. Figure 6 shows the architecture of our proposed discriminator. We use the discriminator to judge the authenticity of each input patch, i.e., the original patch and the reconstructed patch, and update the generator according to the result of the discriminator. By defining $d_w$ as the discriminator and $g_\theta$ as the generator, we can write the loss for discriminator as follows

$$Loss_d = (-1) \times \frac{1}{m} \sum_{i=1}^{m} \left[ d_w\left(S_{(i)}\right) - d_w\left(g_\theta\left(S_{(i)}\right)\right) \right] \quad (8)$$

where $d_w\left(S_{(i)}\right)$ represents the score that $\left(S_{(i)}\right)$ is a ground truth patch, and $d_w\left(g_\theta\left(S_{(i)}\right)\right)$ is the score that the reconstructed patch $g_\theta\left(S_{(i)}\right)$ is a ground truth patch.

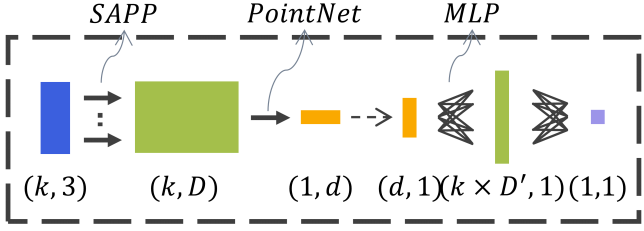

**Figure 6: The architecture of the discriminator network. The input point count received in the discriminator can be either $K$ or $k$.**

And, the loss for generator (i.e, our proposed compression model) now can be rewritten as:

$$Loss_g = D_{cd} + \lambda_1 R + \lambda_2 \underbrace{\frac{1}{m} \sum_{i=1}^{m} \left(-d_w\left(g_\theta\left(S_{(i)}\right)\right)\right)}_{adversarial\ loss} \quad (9)$$

In our experiment, we set $\lambda_1 = 10^{-6}$ and $\lambda_2 = 10^{-3}$.

## 4 EXPERIMENTS

We perform our experiments on ModelNet40 [30], ShapeNet [4] and S3DIS-Area1 (Area 1 in Stanford Large-Scale 3D Indoor Spaces Dataset [3]). The information of data sets and their use purposes are shown in the Tab. 1.

We implement our network by using python 3.9 and pytorch 1.9. For our improved patch-based deep autoencoder (IPDAE), we use Adam optimizer [13] with an initial learning rate of 0.0005 and batch size of 1. For our IPDAE with adversarial learning (IPDAE-GAN), we use RMSprop as the optimizer, and use previously trained IPDAE weights as a starting point for generator. We train 8 epochs for IPDAE and train the generator and discriminator alternatively until convergence for IPDAE-GAN.

### 4.1 Evaluation metric and PreProcessing

We compare our methods with state-of-the-arts, including: previous patch-based deep autoencoder (PDAE) [32], Quach's c1 [22], Quach's c4 [23], PCGCv2 [27], and MPEG G-PCC reference software TMC13 [16]. We use the point-to-point and point-to-plane symmetric PSNR [26, 33] to test the reconstruction quality as in [22], [32]. In addition, we propose a uniformity metric which calculates the variance of the distance of each point to its point cloud. The detailed calculation is described as:

$$dist\,(x, S) = \min_{y \in S} \|x - y\|_2^2 \quad (10)$$

$$SDV\,(S) = \frac{1}{|S|} \sum_{x \in S} \left( dist\,(x, S - \{x\}) \right.$$
$$\left. - \frac{1}{|S|} \sum_{x \in S} dist\,(x, S - \{x\}) \right)^2$$

where $dist$ function defines the distance between a point $x$ and a point cloud $S$, which is the distance between this point and its nearest neighboring point. Self-distance variance (SDV) denotes the distance variance based on the point-to-point distance. Finally, we define uniformity coefficient as the ratio of the uniformity of the reconstructed point cloud to the uniformity of the source point cloud:

$$UC\left(\hat{S}, S\right) = \frac{SDV(\hat{S})}{SDV(S)} \quad (11)$$

A metric expressed as a ratio allows the metric to be size independent and makes the value more intuitive: a UC of 0 represents near-absolute uniformity of the reconstructed point cloud, a UC of 1 represents the similar uniformity as the original point cloud, and a UC greater than 1 represents more inhomogeneity than the origin cloud. The lower the UC value, the more uniform the point set distribution is.

For different methods, the data preprocessing and experimental process are as follows:

For PDAE, we manually zoom in every point cloud coordinates to [0,63] as in [22].

The approach of Quach's c1, Quach's c4 and PCGCv2 are 3D convolution based autoencoders, and their performance depends heavily on the resolution of voxel blocks. Here we zoom in point clouds in ModelNet40 and ShapeNet to [0,63], and [0, 511] for point clouds in S3DIS-Area1. We use the same dataset configuration to test TMC13.

We trained PDAE, Quach's c1 and PCGCv2 using the data set recommended or provided by the authors. For Quach's c4, we use the pretrained network models provided by the authors to test the data set directly. We obtain RD curves of TMC13 by changing position quantization scale.

As for our IPDAE and IPDAE-GAN, we do not need any preprocessing for these three data sets, as we have designed a unified point cloud transformation process and also do not need voxelization. We train and test by applying different $K$ and $d$ for obtaining different compression ratios.

### 4.2 Visualization of Training Process

Figure 1 shows the training process of our IPDAE, where the airplane and chair are from the ModelNet40 test set with 8192 points. The columns from left to right in Fig. 1 represent step 1, 20, 500, 2500, 4000, 10000 and 78500 of our training process respectively. As can be seen, our proposed model is easy to train and can effectively reconstruct complex point cloud shapes in a short time.

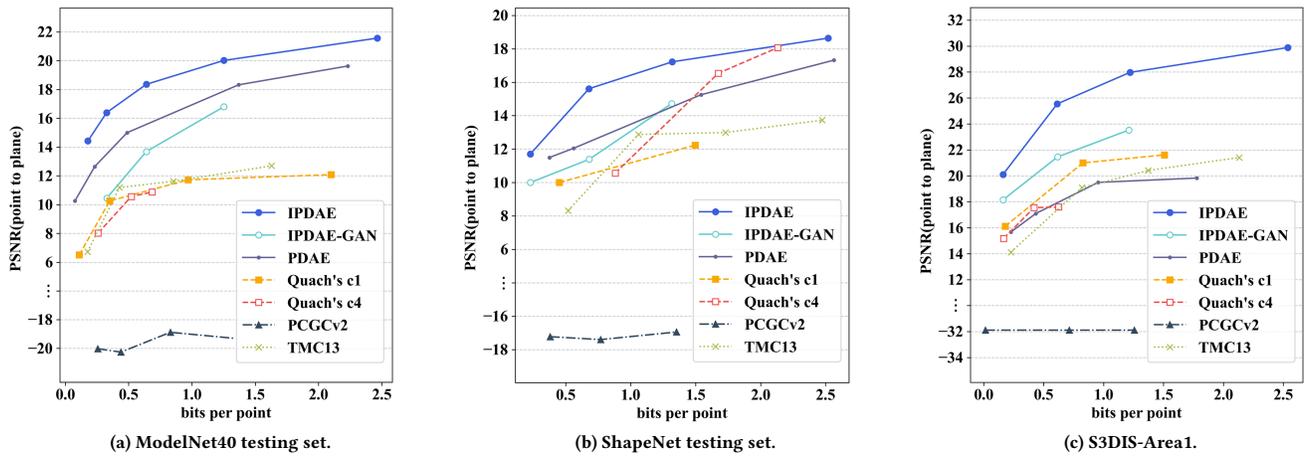

Figure 7: Rate distortion (RD) curves for each dataset.

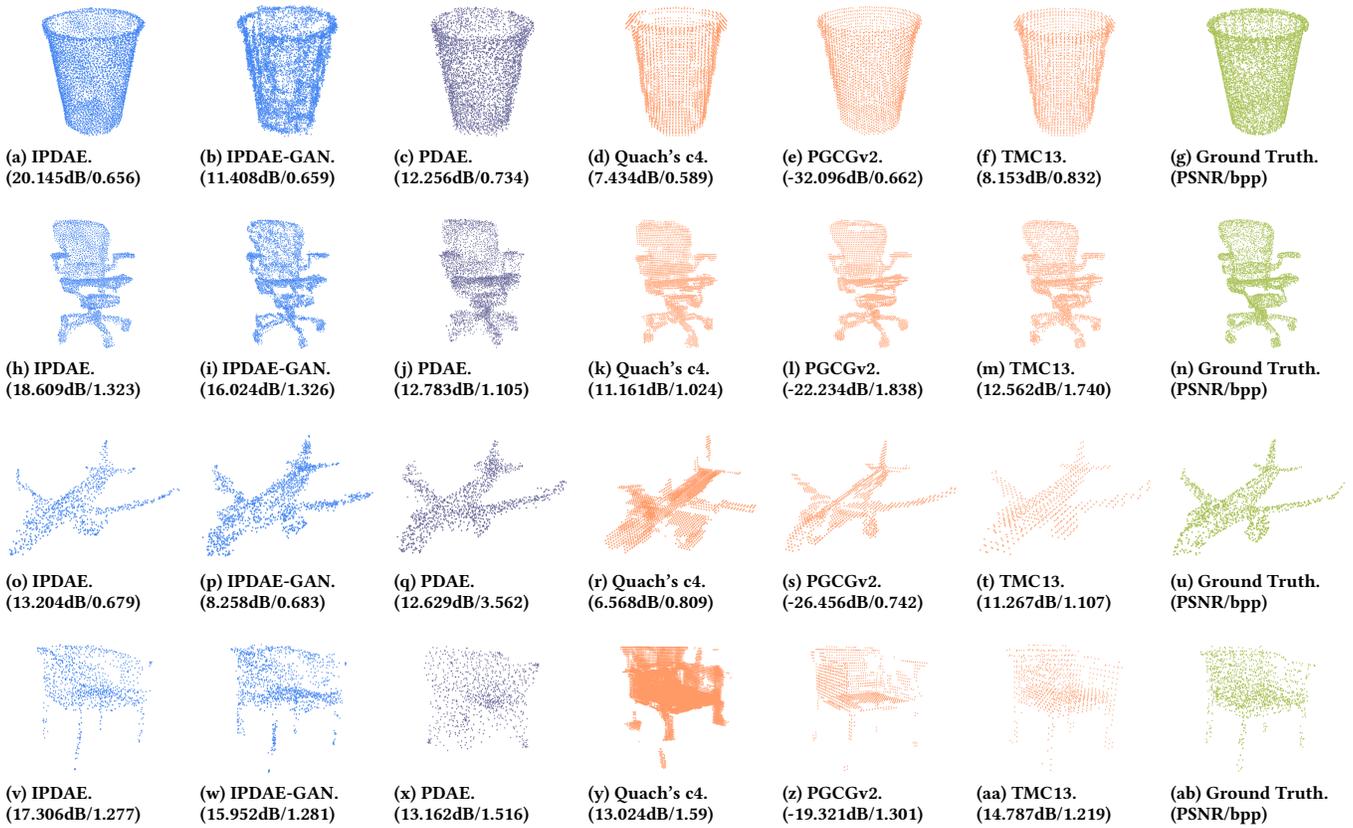

Figure 8: Visual comparison of ompression results. Corresponding point-to-plane PSNR and bpp are shown in brackets. (a) to (n) is the compression results of 8192-points point cloud in ModelNet40, while (o) to (ab) represent the compression results of 2048-points point cloud in ShapeNet.

Table 1: Overview of datasets.

| Dataset | Number of Shapes | Acquisition method | Resolution | Purpose |
| --- | --- | --- | --- | --- |
| ModelNet40 training set | 9843 | Sampling | 8192(8k) | Training |
| ModelNet40 testing set | 2468 | Sampling | 8192(8k) | Validation |
| ShapeNet testing set | 2874 | Sampling | 2048(2k) | Testing |
| S3DIS-Area1 | 44 | Matterport Camera | 0.3M-4M | Testing |

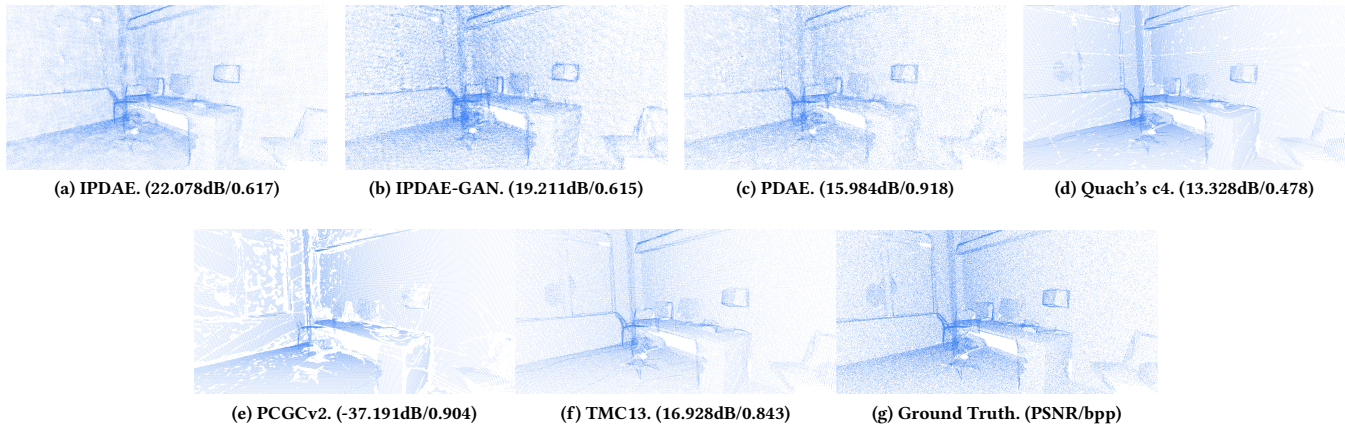

(a) IPDAE. (22.078dB/0.617)   (b) IPDAE-GAN. (19.211dB/0.615)   (c) PDAE. (15.984dB/0.918)   (d) Quach's c4. (13.328dB/0.478)

(e) PCGCv2. (-37.191dB/0.904)   (f) TMC13. (16.928dB/0.843)   (g) Ground Truth. (PSNR/bpp)

Figure 9: Compression of "office_1" in S3DIS-Area1.

## 4.3 Compression performance comparison

Figure 7 shows the compression performance comparison on three datasets. Our IPDAE outperforms other methods on all datasets at all bit rates. For IPDAE-GAN, although equipped with a discriminator, its objective performance is however lower than IPDAE. This could be possibly explained by that the adversarial loss is more focused on visual perception aspect (e.g., the uniformity on the target object) instead of objective distance between two point sets. On the large-scale dataset S3DIS-Area1, our proposed solutions have absolute objective performance advantage compared to others.

Figure 8 presents the visual comparison of the compression results between various methods on ModelNet40 and ShapeNet. As can be observed, at approximately similar bit rate, IPDAE-GAN generally produces visually better reconstructed point cloud. For example, for the chair point cloud in the fourth row, PDAE fails to reconstruct its original shape, and the results of Quach's c4 have obvious block and cluttering effect. In contrast, our proposed solutions have much better visual results, especially for IPDAE-GAN, which reconstructs the intact contour of four legs of the chair. For the "cup" point cloud in the first row, IPDAE generally achieves both better subjective visual results and objective results than IPDAE-GAN.

In Fig. 9, we take the "office_1" point cloud in S3DIS-Area1 as an example to show the compression results of different methods on large-scale dataset. Combining this figure with Fig. 7, we can see that while PDAE performs well on sampled point clouds ShapeNet and ModelNet40, it is overwhelmed by the dense distribution of point clouds in large-scale dataset. This reflects in Fig. 9 (c), where the points in the reconstructed scene is not uniformly distributed, and there are many holes around the table. On the contrary, our transformation and inverse transformation integrated together into an end-to-end compression training process make IPDAE more robust to large-scale point clouds.

From those results, we can also observe that although Quach's c4 is an improved version of c1, however, its performance is not as good as c1 in our experiment, especially on S3DIS dataset. Quach's c4 uses octree to divide the point cloud into several blocks (or cubes), and compress one by one. This may have a good effect on some dense voxel-based point clouds. But it will bring a blocky effect on ordinary non-voxlized point clouds, and also greatly increase the compression/decompression time.

We record the running time and reconstruction uniformity at S3DIS-Area1 of each method in Tab. 2. Each method is tested at a low and a high bit rate scenarios. From the table, our IPDAE can guarantee the optimal reconstruction quality while having acceptable compression/decompression time. In addition, as observed from the results in terms of UC metric, IPDAE-GAN can effectively improve the uniformity of reconstruction points for IPDAE.

## 4.4 Additional Comparison with PCGCv2

As we can see from Fig. 8, the reconstruction quality of PCGCv2 is not that poor visualized by the naked eye, however, the obtained PSNR values are very low, even going to minus. To better understand compression quality of PCGCv2, we further employ Chamfer Distance (CD) to evaluate PCGCv2 and IPDAE.

Because the Chamfer Distance is scale dependent, that is, the distance between points in the coordinate range of [0, 63] is much larger than that in [0, 1]. Therefore, we normalized the outputs of PCGCv2 and IPDAE to [0, 1] for fair comparison.

As shown in Fig. 10, it can be seen that the reconstruction results of our method can better match the original point cloud shape. In

Table 2: Illustration of uniformity and the time spent. bpp refers to bit per point, which is used to represent compression bit rate. D1-PSNR is point-to-point PSNR, and D2-PSNR means point-to-plane PSNR. Tcomp and Tdcomp represent the average time spent on encoding a point cloud in S3DIS-Area1.

| Method | bpp | D1-PSNR | D2-PSNR | UC | Tcomp (s) | Tdcomp (s) | Ttotal (s) |
|---|---|---|---|---|---|---|---|
| IPDAE | 0.161 | 13.161 | 20.109 | 6.409 | 16.38 | 1.21 | 17.59 |
| IPDAE | 0.613 | 16.102 | 25.543 | 3.181 | 57.73 | 3.19 | 60.92 |
| IPDAE-GAN | 0.162 | 12.190 | 18.148 | 4.081 | 16.21 | 1.21 | 17.42 |
| IPDAE-GAN | 0.616 | 13.192 | 21.458 | 2.020 | 58.91 | 3.20 | 62.11 |
| PDAE | 0.227 | 10.488 | 15.675 | 9.531 | 17.55 | 0.06 | 17.61 |
| PDAE | 0.954 | 13.840 | 19.493 | 3.647 | 18.31 | 0.10 | 18.41 |
| Quach's c1 | 0.180 | 7.748 | 16.117 | 0.001 | 0.58 | 127.90 | 128.48 |
| Quach's c1 | 0.827 | 14.796 | 20.995 | 0.000 | 0.56 | 126.45 | 127.01 |
| Quach's c4 | 0.165 | 11.897 | 15.172 | 3.654 | - | - | 605.86 |
| Quach's c4 | 0.420 | 13.528 | 17.547 | 0.028 | - | - | 339.73 |
| PCGCv2 | 0.238 | -42.841 | -31.894 | 0.216 | 0.774 | 0.498 | 1.272 |
| PCGCv2 | 0.714 | -42.841 | -31.894 | 0.206 | 0.809 | 0.507 | 1.316 |
| TMC13 | 0.226 | 9.819 | 14.116 | 0.155 | 16.91 | 2.39 | 19.30 |
| TMC13 | 0.882 | 13.377 | 19.099 | 0.266 | 11.74 | 4.34 | 16.08 |

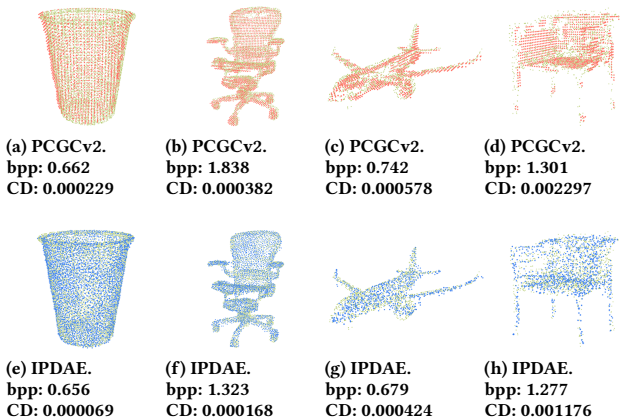

(a) PCGCv2. bpp: 0.662 CD: 0.000229
(b) PCGCv2. bpp: 1.838 CD: 0.000382
(c) PCGCv2. bpp: 0.742 CD: 0.000578
(d) PCGCv2. bpp: 1.301 CD: 0.002297
(e) IPDAE. bpp: 0.656 CD: 0.000069
(f) IPDAE. bpp: 1.323 CD: 0.000168
(g) IPDAE. bpp: 0.679 CD: 0.000424
(h) IPDAE. bpp: 1.277 CD: 0.001176

Figure 10: Comparison of IPDAE and PCGCv2 using Chamfer Distance. The ground truth and reconstrcuted point clouds are superimposed, where green points represent the Ground Truth, and, red points and blue points represent PCGCv2 and IPDAE reconstruction results respectively. (Best viewed in color and zoom in.)

addition to the offset errors caused by voxelization, PCGCv2 will also induce many holes in reconstruction. This regional absence will have a great impact on both CD and PSNR.

### 4.5 Ablation Study

In this section, we present ablation experiments to explore the impact of each component on compression performance.

As shown in Fig. 11, global loss greatly improves compression efficiency, which undoubtedly uncovers one of the key aspects of patch-based reconstruction that is not well-considered in [32]. In addition, we can observe that the octree coding of sampled points can effectively reduce the bitrate while maintaining the reconstruction quality loss relatively small. The proposed learnable context model also contributes a certain amount of performance improvement to learned point cloud geometry compression.

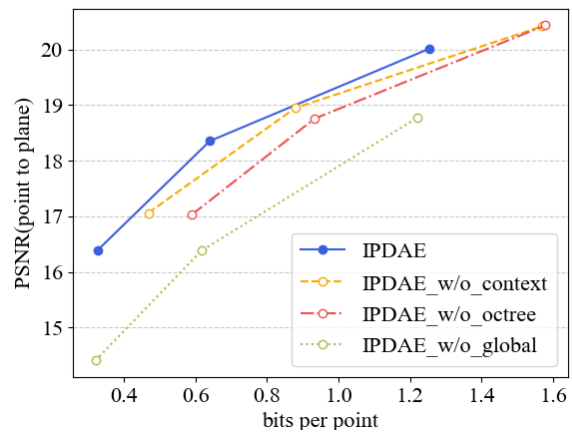

Figure 11: Ablation study on ModelNet40. IPDAE_w/o_context means removal of learnable context model for entropy coding, w/o_octree means removal of octree coding for sampling centroid points, w/o_global means using patch-to-patch criterion instead of global criterion.

## 5 CONCLUSION

In this paper, we propose a deep learning based lossy geometry compression framework, which makes a series of improvements for patch-based autoencoder, including octree coding for sampling points, learnable context model for entropy coding, and integrated compression and training process. Like learned image and video coders in the literature, our improved patch-based autoencoder can be end-to-end trained, which boosts the performance of point cloud compression significantly. Our proposed compression is also generic, which can work effectively on both on sparse or large-scale point clouds. Compared to 3D convolution based solutions, our proposed network not only brings significant compression performance improvement, but also is time efficient. Finally, we also include adversarial learning to improve the uniformity of point cloud reconstruction. The implementation code and pretrained models will be made publicly available.

This supplementary material provides specific network architectures of the proposed IPDAE and IPDAE models used for experiments in the main paper. In addition, we introduce an extended mode for IPDAE to compress automotive LiDAR point clouds.

# 1 SPECIFIC PARAMETER SETTINGS

## 1.1 IPDAE

In all our experiments, we set:

| Param | Value | Description |
|---|---|---|
| $n_0$ | 1024 | Reference Quantity Constant |
| $\alpha$ | 2 | Patch Coverage |
| $L$ | 7 | Quantization Level |
| $d$ | 16 | AE Bottleneck Size |
| $\lambda$ | $10^{-6}$ | Balance of $D_{cd}$ and $R$ |

Table 1: Hyper parameters.

We achieve compression at different bit rates mainly by changing the resolution of each patch ($K$) and bit rate of octree coded bitstream ($R_{oc}$). The following shows the two parameter selections for obtaining different rate distortion points on ModelNet40 testing dataset. We use the same $K$ and $R_{oc}$ configuration for other datasets.

| $K$ | $R_{oc}$ | bpp | D2-PSNR |
|---|---|---|---|
| 1024 | 0.07 | 0.178 | 14.416 |
| 512 | 0.125 | 0.326 | 16.389 |
| 256 | 0.25 | 0.640 | 18.354 |
| 128 | 0.5 | 1.253 | 20.013 |
| 64 | 1.0 | 2.464 | 21.561 |

Table 2: Compression at different bit rates.

In addition, for a specific point cloud with $n$ points, we can get the number of patches $m = \lfloor \frac{\alpha n}{K} \rfloor$. The resolution of a reconstructed patch is set to $k = \lfloor \frac{K}{\alpha} \rfloor$. In this setting, the resolution of the reconstructed whole point cloud by our network is $m \times k$ as described in the main paper.

For the neural network architecture, specifically, we use the following parameters.

- Context entropy model:
$$PointNet(SmlpSize = [3, 64, 128, 256]) \rightarrow SMLP([3 + 256, 256, 512, 16 \times 7]) \rightarrow Reshape(16, 7)$$
- Encoder of improved autoencoder:
$$SAPP(GroupSize = 16, SmlpSize = [3, 32, 64, 128]) \rightarrow PointNet([3 + 128, 128, 256, 512, 16])$$
- Decoder of improved autoencoder:
$$MLP([16, 256, 1024, k \times 128]) \rightarrow Reshape(k, 128) \rightarrow SMLP([16 + 128, 128, 64, 32, 3])$$

Where MLP is an abbreviation for multilayer perceptron and SMLP means Shared-MLP. SAPP means set abstraction per point, similar to set abstraction layer in PointNet++.

## 1.2 IPDAE-GAN

We use the IPDAE structure with the same parameters as the generator for our experiments. For the discriminator network architecture, specifically, we use the following parameters:
$$SAPP(GroupSize = 16, SmlpSize = [3, 32, 64, 128]) \rightarrow PointNet([3 + 128, 128, 256, 512, 16]) \rightarrow MLP([16, 256, k \times 128, 128, 1])$$

# 2 COMPRESSION RESULTS ON SEMANTICKITTI

The point cloud with uneven point distribution has always been difficult to be processed by PointNet structure, especially in the task of reconstruction. Point clouds in SemanticKITTI dataset are captured by the dedicated automotive LiDAR, which will make the points very dense in the area close to the LiDAR but quite sparse in the place far away.

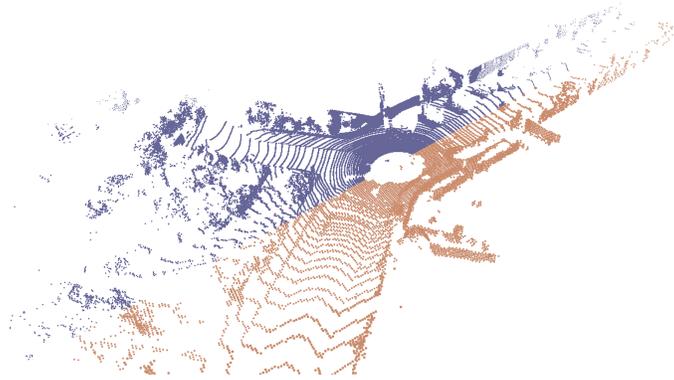

Figure 1: Visualization of SematicKITTI voxelization results. Original point cloud is presented in blue, and the associated point cloud after voxelization is illustrated in brown. These two point clouds were spliced with half from each for a clearer contrast.

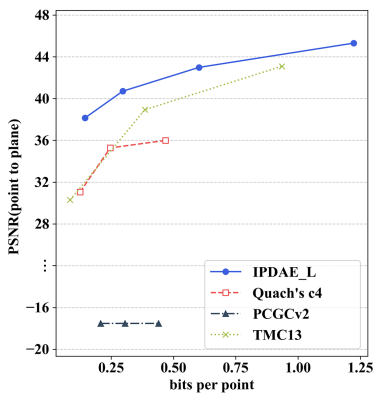

Figure 2: Rate distortion (RD) curves for SemanticKITTI. IPDAE_L represents the extended mode of IPDAE.

Figure. 1 shows voxelization results on SematicKITTI dataset. As we can observe from Fig. 1, because of the limited voxel resolution, voxelization can reduce the unevenness to a certain extent. Points which are too close will be grouped into one point/voxel. In a word, voxelization sacrifices precision in exchange for uniformity.

However, although the patch-based method takes the original set as the input directly, too-dense points will lead to the failure of reconstruction. In order to make our model better handle this kind of data, we design an extended mode for IPDAE, which contains two main changes:

Firstly, we use random point sampling with $\alpha = 6$ instead of farthest point sampling with $\alpha = 2$. This setting can better cover the uneven point set in SematicKITTI.

Secondly, we use patch scale normalization operation instead of simple rescale at Eq. (1) in the main paper. Specifically, each patch is rescaled to the unit sphere $[0, 1]$. The original scale information of patches are transmitted to the decoder to restore the original size. Since each patch includes the same number of points, this patch scale normalization operation allows each patch to maintain the same density as the training data.

Figure 2 shows the performance comparison on SemanticKITTI. We use the same preprocess configuration as S3DIS-Area1 in the main text. Our method substantially outperforms the MPEG reference solution TMC13 in terms of rate-distortion performance, with an average of 46.64% BDBR savings.

In Fig. 3, we show two examples of compression qualitative results in the first sequence of SemanticKITTI. As can be seen, our method can guarantee to generate the same number of points as the input, while keeping the original density distribution. Note that, TMC13 is a traditional compression platform, which processes the point cloud based on hand-crafted features

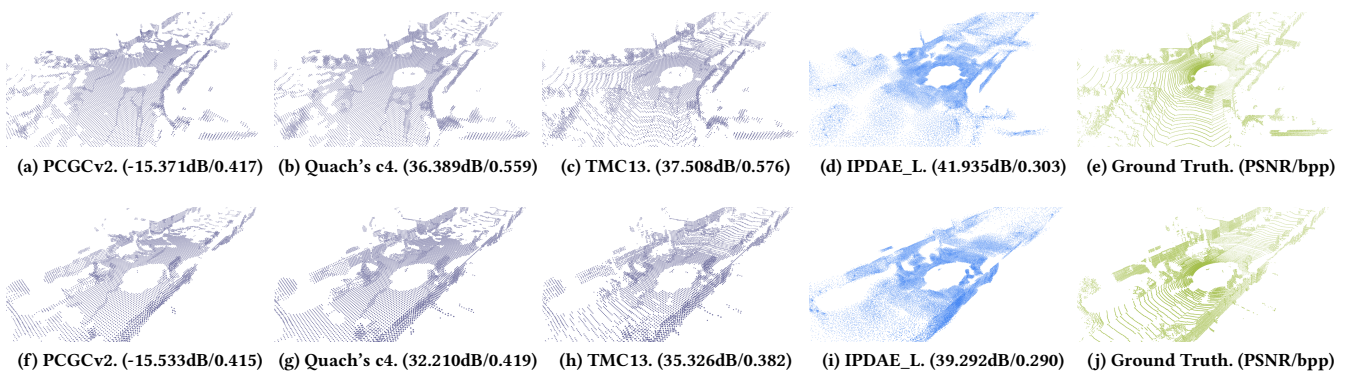

(a) PCGCv2. (-15.371dB/0.417)  (b) Quach's c4. (36.389dB/0.559)  (c) TMC13. (37.508dB/0.576)  (d) IPDAE_L. (41.935dB/0.303)  (e) Ground Truth. (PSNR/bpp)

(f) PCGCv2. (-15.533dB/0.415)  (g) Quach's c4. (32.210dB/0.419)  (h) TMC13. (35.326dB/0.382)  (i) IPDAE_L. (39.292dB/0.290)  (j) Ground Truth. (PSNR/bpp)

Figure 3: Compression of frame "000000" and "000042" in the sequence "00".

without concerning the density or sparsity, and it generally can better preserve the structure (e.g., the contour lines) of the point clouds in SemanticKITTI compared to learned solutions.

## 3 ADDITIONAL DESCRIPTION OF TMC13 PARAMETERS

In all of our experiments, we use TMC13 version v14, and set its configuration parameters following common test conditions as follows:

Table 3: Encoder Settings.

| Parameter | Value |
| --- | --- |
| partitionMethod | 2 |
| partitionOctreeDepth | 10 |
| seq_bounding_box_whd | 0, 0, 0 |
| positionQuantizationScale | - |
| positionQuantisationOctreeDepth | 10 |
| positionQuantisationEnabled | 1 |
| positionBaseQp | 8 |
| qp | 5 |
| bitdepth | 10 |
| transformType | 1 |
| rahtDepth | 10 |
| levelOfDetailCount | 12 |

Table 4: Decoder Settings.

| Parameter | Value |
| --- | --- |
| mode | 1 |
| compressedStreamPath | - |
| reconstructedDataPath | - |
| skipOctreeLayers | 0 |

We change the compression ratio by setting positionQuantizationScale to a value between 0.0 and 1.0.